\documentclass[conference]{IEEEtran}
\usepackage[letterpaper, top=0.75in, bottom=0.95in, left=0.63in, right=0.63in]{geometry}
\usepackage[nolist]{acronym}
\usepackage{fontawesome}
\usepackage{amsmath}
\usepackage{tikz}

\usepackage{graphicx}
\usepackage{subcaption}
\usepackage{xcolor}     % Para cores (mais moderno que o color)
\usepackage{tabularray}  % OBRIGATÓRIO para o ambiente tblr funcionar
\let\labelindent\relax
\usepackage{enumitem}
\usepackage[T1]{fontenc}
\usepackage{url}
\usepackage{booktabs}
\usepackage{multirow}
\usetikzlibrary{arrows.meta,positioning,calc,shadows}
\usepackage{algorithm}
\usepackage{algorithmic}
\usepackage{booktabs}
\usepackage{textcomp}
\usepackage{manyfoot}
\usepackage{listings}
\usepackage{url}
\usepackage{comment}
\usepackage{tikz}
\usepackage{helvet}
\usepackage{subcaption}
\usetikzlibrary{arrows.meta, shadows.blur}
\usepackage{amsfonts}
\definecolor{HeaderSoftBlue}{HTML}{D9EAF7}
\definecolor{OurSoftGray}{HTML}{F2F2F2}
\usepackage{placeins}
\definecolor{cF}{HTML}{1565c0}   % Phase 1 – Fabric blue
\definecolor{cL}{HTML}{e65100}   % Phase 2 – LoRA orange
\definecolor{cM}{HTML}{2e7d32}   % Phase 3 – Monitor green
\definecolor{cP}{HTML}{4527a0}   % Phase 4 – Perf purple
\definecolor{cA}{HTML}{b71c1c}   % Phase 5 – ANOVA red
\definecolor{cC}{HTML}{6a1b9a}   % Chameleon
\definecolor{cGr}{HTML}{eceff1}  % light grey fill
\definecolor{cDk}{HTML}{263238}  % dark text

\tikzset{
  MA/.style={-{Latex[length=7pt,width=4.5pt]}, line width=2pt,   draw=cDk},
  SA/.style={-{Latex[length=5pt,width=3.5pt]}, line width=1.2pt, draw=cDk!65},
  DA/.style={-{Latex[length=5pt,width=3pt]},   line width=1.1pt, draw=cDk!50, dashed},
}

\ifCLASSINFOpdf
\else
\fi

\usepackage{stfloats}
\begin{document}
\acrodef{AI}{Artificial Intelligence}
\acrodef{ALL}{Acute Lymphoblastic Leukemia}
\acrodef{ANOVA}{Analysis of Variance}
\acrodef{AIaaS}{AI-as-a-Service}
\acrodef{CAISHI}{Cervical Adenocarcinoma in Situ}
\acrodef{CT}{Computed Tomography}
\acrodef{CNN}{Convolutional Neural Network}
\acrodef{CRC}{Colorectal cancer}
\acrodef{CNNs}{Convolutional Neural Networks}
\acrodef{CCT}{Compact Convolutional Transformer}
\acrodef{CPU}{Central Processing Unit}
\acrodef{CLM}{Compact Language Models}
\acrodef{DT}{Decision Tree}
\acrodef{DEPO}{Dynamic Energy‑Performance Optimizer}
\acrodef{ECBAM}{Enhanced Convolutional Block Attention Modules}
\acrodef{FL}{Federated Learning}
\acrodef{FedAvg}{Federated Averaging}
\acrodef{HCPA}{Hospital de Clínicas de Porto Alegre}
\acrodef{HIPAA}{Health Insurance Portability and Accountability Act}
\acrodef{IoT}{Internet of Things}
\acrodef{GA}{Genetic Algorithm}
\acrodef{GDPR}{General Data Protection Regulation}
\acrodef{GWO}{Grey Wolf Optimizer}
\acrodef{GPU}{Graphic Processing Unit}
\acrodef{GPUs}{Graphic Processing Units}
\acrodef{HPC}{High-Performance Computing}
\acrodef{KNN}{K-Nearest Neighbors}
\acrodef{non-IID}{non-independent and identically distributed}
\acrodef{LoRA}{Low-Rank Adaptation}
\acrodef{LLM}{Large Language Model}
\acrodef{LLMs}{Large Language Models}
\acrodef{MSD}{Micro-Structure Descriptor}
\acrodef{PCA}{Principal Component Analysis}
\acrodef{PSO}{Particle Swarm Optimization}
\acrodef{ProxyNCA}{Proxy Neighborhood Component Analysis}
\acrodef{PEFT}{Parameter-Efficient Fine-Tuning}
\acrodef{QLoRA}{Quantized Low-Rank Adaptation}
\acrodef{RF}{Random Forest}
\acrodef{RL}{Reinforcement Learning}
\acrodef{RWKV}{Receptance Weighted Key Value}
\acrodef{SVM}{Support Vector Machine}
\acrodef{SVD}{Singular Value Decomposition}
\acrodef{ViT}{Vision Transformer}

%\title{Forecasting User Experience on Mobile Networks through Infrastructure Monitoring in container-based environments}
%\title{Forecasting User Experience in Mobile Networks through Generic Monitoring-Driven Approaches}
%\title{Leveraging Infrastructure Monitoring for User Experience Forecasting in container-based \\ Mobile Network Core}
%\title{Leveraging Infrastructure Monitoring for User Experience Forecasting in Container-based 5G Core}

%\title{Using Post-Quantum Cryptography to Enhance Resilience Against Cyberattacks in 5G and Future Networks}

%\title{Energy Aware Agentic Orchestration for Geographically Distributed Service Deployment}
\title{Sustainable Federated Fine-Tuning with LoRA: Environmental Impact and Performance Footprint}

\author{ \IEEEauthorblockN{ Pedro Augusto Londe Ribeiro\textsuperscript{1}, Rodrigo Moreira\textsuperscript{1}, Larissa Ferreira Rodrigues Moreira\textsuperscript{1},\\ Joberto S. B. Martins\textsuperscript{2}, Tereza C. Carvalho\textsuperscript{3}, Flávio de Oliveira Silva\textsuperscript{4} } \IEEEauthorblockA{ \textsuperscript{1}Federal University of Viçosa (UFV), Minas Gerais, Brazil\\ Emails: \{pedro.londe.ribeiro, rodrigo, larissa.f.rodrigues\}@ufv.br } \IEEEauthorblockA{ \textsuperscript{2}University of Salvador (UNIFACS), Salvador, Brazil\\ Email: joberto@ieee.org } \IEEEauthorblockA{ \textsuperscript{3}University of São Paulo (USP), São Paulo, Brazil\\ Email: terezacarvalho@usp.br } \IEEEauthorblockA{ \textsuperscript{4}University of Minho (UMinho), Department of Informatics, ALGORITMI Centre, Braga, Portugal\\ Email: flavio@di.uminho.pt } \thanks{Rodrigo Moreira (ORCID: 0000-0002-9328-8618) is the corresponding author.} }

\maketitle

\begingroup
\renewcommand\thefootnote{}
\footnotetext{\scriptsize
© 2026 IEEE. Personal use of this material is permitted. Permission from IEEE must be obtained for all other uses, in any current or future media, including reprinting/republishing this material for advertising or promotional purposes, creating new collective works, for resale or redistribution to servers or lists, or reuse of any copyrighted component of this work in other works.

Accepted author manuscript for publication in the Proceedings of the 2027 IEEE Consumer Communications \& Networking Conference (CCNC 2027), Las Vegas, Nevada, USA, January 9--12, 2027.
}
\addtocounter{footnote}{-1}
\endgroup

\begin{abstract}
Federated Learning (FL) with Parameter-Efficient Fine-Tuning (PEFT) enables efficient adaptation of Large Language Models (LLMs), yet its energy and carbon costs across hardware remain poorly understood. We evaluated federated Low-Rank Adaptation (LoRA) fine-tuning on NVIDIA A40, RTX 6000 Ada, and A100 GPUs using a $2^2$ full factorial design \textcolor{black}{with 135M/360M models and 10/100 local steps}. \textcolor{black}{CodeCarbon tracked energy and estimated CO$_2$-equivalent emissions. Descriptive allocation of variation assigned} 84.8–91.2\% \textcolor{black}{of energy and emissions variation to local steps within these ranges}. \textcolor{black}{On A40 and RTX 6000 Ada,} configurations with fewer steps achieved comparable accuracy with \textcolor{black}{roughly} 6–8-fold lower energy consumption and higher accuracy per kWh. \textcolor{black}{These findings support smaller local budgets for energy-efficient federated fine-tuning within the evaluated setting.}
\end{abstract}

\begin{IEEEkeywords} Federated Learning, Large Language Models, Parameter-Efficient Fine-Tuning (PEFT), Energy Consumption, Environmental Sustainability \end{IEEEkeywords}

\IEEEpeerreviewmaketitle

\section{Introduction}\label{sec:introduction}

Training and using \acp{LLM} is quickly becoming a major source of electricity demand in computing. The International Energy Agency (IEA) predicts that data center energy use will more than double by 2030, reaching approximately 945\,TWh, mainly owing to \ac{AI} tasks~\cite{su18031359, Alexandre2026}. Industry estimates from Goldman Sachs suggest a 160\% rise in data-center power demand by the end of the decade, possibly accounting for up to 4\% of global electricity use~\cite{goldmansachs2024_ai_power_demand}. These predictions lead to a clear question: how much energy does each part of the modern \ac{LLM} process use, and where can we reduce it the most?

\ac{FL}, a privacy-preserving learning paradigm, combined with \ac{PEFT} techniques like \ac{LoRA} provides a practical way to adapt \ac{LLMs} to specific tasks while keeping data decentralized and reducing the number of trainable parameters~\cite{Wu2024, Hinov2025}. Additional strategies, such as probability-level ensembling that shares lightweight outputs instead of model weights, further cut communication costs~\cite{Moreira2026}. However, research on these methods has mainly focused on accuracy, communication efficiency, and privacy guarantees. What is still mostly unmeasured is the energy and carbon cost of federated fine-tuning when it operates on diverse hardware, which is typical in real deployments with different \ac{GPU} generations, thermal design powers, and memory subsystems across federated clients and cloud platforms~\cite{1_Iftikhar2025, 13_Herbert2024}.

\textcolor{black}{Our paper examines whether local optimization intensity outweighs model size in energy use and CO$_2$-equivalent emissions within the tested ranges. We evaluated NVIDIA A40, RTX 6000 Ada, and A100 platforms separately.} Our contributions are threefold: (i) we measured energy consumption and \textcolor{black}{estimated} CO$_2$-equivalent emissions during federated \ac{LLM} fine-tuning across three \ac{GPU} families \textcolor{black}{using CodeCarbon and GPU telemetry}; (ii) we used a $2^2$ factorial design \textcolor{black}{to allocate observed variation among} model size, local optimization steps, and their interaction, \textcolor{black}{with local steps accounting for 84.8--91.2\% of energy and emissions variation within the selected levels}; (iii) \textcolor{black}{we related predictive performance to energy cost through accuracy per kWh}.

The remainder of this paper is structured as follows.  Section~\ref{sec:related_work} reviews related work on sustainable \ac{LLM} adaptation and energy accounting.  Section~\ref{sec:method} presents the experimental methodology.  Section~\ref{sec:results_and_discussion} reports and discusses the results.  Section~\ref{sec:concluding_remarks} concludes with limitations and directions for future work.

\section{Related Work}\label{sec:related_work}

This section reviews recent work on sustainable \ac{LLM} adaptation, organized into three thematic groups: federated and edge fine-tuning with \ac{PEFT}, measurement-driven energy and carbon accounting on \acp{GPU}, and systems or benchmarking efforts that optimize or standardize evaluation.

\textbf{Federated and edge fine-tuning with PEFT}. Iftikhar et al.~\cite{1_Iftikhar2025} study energy-aware client selection for federated \ac{LoRA}, reporting accuracy and CO$_2$e estimates. Wu et al.~\cite{4_Wu2026} use truncated \ac{SVD} with dynamic rank allocation to improve accuracy and communication efficiency in heterogeneous federations. Thuau et al.~\cite{7_Thuau2025} compare federated \ac{LoRA} with compact personalized models for violence detection, including energy and calibration analyses. \textcolor{black}{Our $2^2$ factorial design allocates observed energy and emissions variation to model size, local steps, and their interaction within the tested ranges.}

On edge devices, Xu et al.~\cite{10_Xu2024} combine perturbed forward passes with \ac{LoRA} to reduce memory requirements. Woisetschl\"{a}ger et al.~\cite{13_Herbert2024} evaluate federated T5 fine-tuning at the extreme edge using a real-time energy-efficiency metric. Li et al.~\cite{2_Li2025} combine split learning and \ac{LoRA} to reduce latency and server energy under privacy constraints, while Zhao et al.~\cite{12_Zhao2026} jointly optimize communication and computation in unmanned vehicle networks. \textcolor{black}{We focus on software-based energy accounting during simulated federated fine-tuning.}

\textbf{Energy and carbon accounting on GPUs}. Krzywaniak et al.~\cite{9_Krzywaniak2023} introduce dynamic power capping with online performance tracing \ac{DEPO} for \ac{HPC} workloads. Newkirk et al.~\cite{11_Newkirk2025} use measured power traces and statistical models to relate workload intensity to node-level energy. Gogineni et al.~\cite{15_Gogineni2025} examine single-\ac{GPU} fine-tuning bottlenecks involving \ac{LoRA}, checkpointing, and power capping. We compare three \ac{GPU} families under the same training configurations.

At the system level, Ozkan and Ozkan~\cite{14_Ozkan2025} combine telemetry and grid carbon intensity to schedule inference across heterogeneous fleets. Li et al.~\cite{5_Li2025} use \ac{LoRA} with a \ac{RWKV} model to predict product carbon footprints. Instead, our study quantifies the energy consumption and CO$_2$-equivalent emissions associated with federated fine-tuning.

\textbf{Benchmarking and efficiency surveys}. Kuang et al.~\cite{6_Kuang2024} provide FederatedScope~\ac{LLM}, a reproducible framework with standardized \ac{PEFT} implementations and configurable evaluation. \textcolor{black}{Hanindhito} et al.~\cite{3_Hanindhito2025} analyze the compute and memory trade-offs of \ac{LoRA} and \ac{QLoRA} across configurations and \acp{GPU}. Kaushik et al.~\cite{8_Kaushik2025} survey sustainability challenges and mitigation strategies, including pruning, quantization, and sparsity. We complement these efforts with an accuracy-per-kWh metric that relates predictive performance to energy cost.

\begin{table}[!t] \centering \caption{Comparison with related work.} \label{tab:rw_energy_fedllm_matrix} \scriptsize \setlength{\tabcolsep}{2.5pt} \renewcommand{\arraystretch}{1.05} \begin{tabular}{lcccccc} \hline \textbf{Paper} & \textbf{Federated} & \textbf{PEFT} & \shortstack{\textbf{Direct}\\\textbf{energy}} & \shortstack{\textbf{Carbon}\\\textbf{CO$_2$e}} & \shortstack{\textbf{GPU}\\\textbf{comparison}} & \shortstack{\textbf{Factorial}\\\textbf{analysis}} \\ \hline Iftikhar~\cite{1_Iftikhar2025} & \faCircle & \faCircle & \faCircle & \faCircle & \faCircleO & \faCircleO \\ Li~\cite{2_Li2025} & \faCircleO & \faCircle & \faCircleO & \faCircleO & \faCircleO & \faCircleO \\ Hanindhito~\cite{3_Hanindhito2025} & \faCircleO & \faCircle & \faCircleO & \faCircle & \faCircle & \faCircleO \\ Wu~\cite{4_Wu2026} & \faCircle & \faCircle & \faCircleO & \faCircleO & \faCircleO & \faCircleO \\ Li~\cite{5_Li2025} & \faCircleO & \faCircle & \faCircleO & \faCircle & \faCircleO & \faCircleO \\ Kuang~\cite{6_Kuang2024} & \faCircle & \faCircle & \faCircleO & \faCircleO & \faCircle & \faCircleO \\ Thuau~\cite{7_Thuau2025} & \faCircle & \faCircle & \faCircle & \faCircle & \faCircleO & \faCircleO \\ Kaushik~\cite{8_Kaushik2025} & \faCircleO & \faCircle & \faCircleO & \faCircleO & \faCircleO & \faCircleO \\ Krzywaniak~\cite{9_Krzywaniak2023} & \faCircleO & \faCircleO & \faCircle & \faCircle & \faCircle & \faCircleO \\ Xu~\cite{10_Xu2024} & \faCircle & \faCircle & \faCircleO & \faCircleO & \faCircle & \faCircleO \\ Newkirk~\cite{11_Newkirk2025} & \faCircleO & \faCircleO & \faCircle & \faCircle & \faCircle & \faCircleO \\ Zhao~\cite{12_Zhao2026} & \faCircle & \faCircle & \faCircleO & \faCircleO & \faCircleO & \faCircleO \\ Woisetschl\"ager~\cite{13_Herbert2024} & \faCircle & \faCircle & \faCircleO & \faCircleO & \faCircle & \faCircleO \\ Ozkan~\cite{14_Ozkan2025} & \faCircle & \faCircleO & \faCircleO & \faCircle & \faCircle & \faCircleO \\ Gogineni~\cite{15_Gogineni2025} & \faCircleO & \faCircle & \faCircleO & \faCircleO & \faCircle & \faCircleO \\ \hline \textbf{Our approach} & \faCircle & \faCircle & \faCircle & \faCircle & \faCircle & \faCircle \\ \hline \end{tabular} \end{table}

Table~\ref{tab:rw_energy_fedllm_matrix} summarizes six methodological and reporting features; \faCircle{} denotes a reported feature and \faCircleO{} a feature not reported. ``PEFT'' includes \ac{LoRA}, ``Direct energy'' denotes instrumented measurements, and ``Carbon'' denotes reported CO$_2$e estimates. ``GPU comparison'' indicates evaluation across distinct accelerator families. ``Factorial analysis'' denotes the separation of main effects and interactions on energy or emissions through a factorial design. \textcolor{black}{Our study combines platform measurements with descriptive allocation of variation.}

\section{Energy-efficient Evaluation Method}\label{sec:method}

\begin{figure*}[ht]
    \centering
    \resizebox{\textwidth}{!}{%
        \begin{tikzpicture}[
            font=\sffamily,
            SA/.style={-{Latex[length=4pt]}, line width=0.6pt, draw=black!70},
            MA/.style={-{Latex[length=6pt]}, line width=1.2pt, draw=cDk!60},
            DA/.style={-{Latex[length=6pt]}, dashed, line width=1pt, draw=cDk!50}
        ]

        \draw[draw=cF!45, fill=cF!7,  rounded corners=6pt, dashed, line width=0.9pt]
            (-2.8, 1.8) rectangle (3.3, 6.6);
        \draw[draw=cL!45, fill=cL!6,  rounded corners=6pt, dashed, line width=0.9pt]
            (3.7,  1.8) rectangle (11.4, 6.6);
        \draw[draw=cM!45, fill=cM!6,  rounded corners=6pt, dashed, line width=0.9pt]
            (11.8, 1.8) rectangle (19.6, 6.6);
        \draw[draw=cP!45, fill=cP!6,  rounded corners=6pt, dashed, line width=0.9pt]
            (20.0, 1.8) rectangle (26.0, 6.6);
        \draw[draw=cA!45, fill=cA!5,  rounded corners=6pt, dashed, line width=0.9pt]
            (-2.8, -4.2) rectangle (26.0, 1.2);

        \node[draw=cF!70, fill=cF!15, rounded corners=4pt,
              minimum width=5.6cm, minimum height=1.1cm, align=center,
              blur shadow={shadow blur steps=4,shadow xshift=1pt,shadow yshift=-1pt}]
            at (0.25, 6.0) {\textbf{Phase 1}\\[1pt]Testbed Infrastructure};

        \node[draw=cF!60, fill=cF!12, rounded corners=3pt,
              minimum width=2.3cm, minimum height=0.9cm, align=center,
              font=\sffamily\normalsize] (fab)
            at (-1.1, 4.6) {\textbf{FABRIC}};

        \node[draw=cC!60, fill=cC!10, rounded corners=3pt,
              minimum width=2.3cm, minimum height=0.9cm, align=center,
              font=\sffamily\normalsize] (cham)
            at ( 1.6, 4.6) {\textbf{Chameleon}};

        \node[draw=cF!50, fill=cF!8, rounded corners=2pt,
              minimum width=1.7cm, minimum height=0.7cm, align=center,
              font=\sffamily\normalsize] (a100) at (-1.9, 3.1) {A40};

        \node[draw=cF!50, fill=cF!8, rounded corners=2pt,
              minimum width=1.7cm, minimum height=0.7cm, align=center,
              font=\sffamily\normalsize] (a40)  at ( 0.2, 3.1) {RTX6000};

        \node[draw=cC!50, fill=cC!8, rounded corners=2pt,
              minimum width=1.7cm, minimum height=0.7cm, align=center,
              font=\sffamily\normalsize] (rtx)  at ( 2.3, 3.1) {A100};

        \draw[SA] (fab.south)  -- (a100.north);
        \draw[SA] (fab.south)  -- (a40.north);
        \draw[SA] (cham.south) -- (rtx.north);

        \node[draw=cL!70, fill=cL!15, rounded corners=4pt,
              minimum width=7.0cm, minimum height=1.1cm, align=center,
              blur shadow={shadow blur steps=4,shadow xshift=1pt,shadow yshift=-1pt}]
            at (7.55, 6.0) {\textbf{Phase 2}\\[1pt]Experimental Design};

        \node[draw=cL!60, fill=cL!8, rounded corners=3pt,
              minimum width=6.8cm, minimum height=1.55cm,
              align=center, inner sep=6pt, font=\sffamily\small]
            (design) at (7.55, 4.55) {
            \textbf{$2^2$ \textcolor{black}{Full} Factorial Design}\\[3pt]
        };

        \node[draw=cL!45, fill=cL!6, rounded corners=2pt,
              minimum width=1.55cm, minimum height=0.72cm,
              align=center, font=\sffamily\scriptsize] at (4.85, 2.95) {135M\\10 steps};
        \node[draw=cL!45, fill=cL!6, rounded corners=2pt,
              minimum width=1.55cm, minimum height=0.72cm,
              align=center, font=\sffamily\scriptsize] at (6.45, 2.95) {135M\\100 steps};
        \node[draw=cL!45, fill=cL!6, rounded corners=2pt,
              minimum width=1.55cm, minimum height=0.72cm,
              align=center, font=\sffamily\scriptsize] (e3) at (8.05, 2.95) {360M\\10 steps};
        \node[draw=cL!45, fill=cL!6, rounded corners=2pt,
              minimum width=1.55cm, minimum height=0.72cm,
              align=center, font=\sffamily\scriptsize] at (9.65, 2.95) {360M\\100 steps};

        \draw[SA] (design.south) -- ++(0,-0.26) -| (4.85, 2.95+0.36);
        \draw[SA] (design.south) -- ++(0,-0.26) -| (6.45, 2.95+0.36);
        \draw[SA] (design.south) -- ++(0,-0.26) -| (8.05, 2.95+0.36);
        \draw[SA] (design.south) -- ++(0,-0.26) -| (9.65, 2.95+0.36);

        \node[draw=cM!70, fill=cM!15, rounded corners=4pt,
              minimum width=7.0cm, minimum height=1.1cm, align=center,
              blur shadow={shadow blur steps=4,shadow xshift=1pt,shadow yshift=-1pt}]
            at (15.7, 6.0) {\textbf{Phase 3}\\[1pt]Energy Monitoring};

        \node[draw=cM!60, fill=cM!12, rounded corners=3pt,
              minimum width=6.6cm, minimum height=0.95cm,
              align=center, inner sep=5pt] (cc)
            at (15.7, 4.6) {\textbf{CodeCarbon}};

        \node[draw=cM!50, fill=cM!7, rounded corners=2pt,
              minimum width=1.4cm, minimum height=0.7cm,
              align=center, font=\sffamily\normalsize] at (13.0, 3.1) {GPU\\Power (W)};
        \node[draw=cM!50, fill=cM!7, rounded corners=2pt,
              minimum width=1.4cm, minimum height=0.7cm,
              align=center, font=\sffamily\normalsize] at (14.7, 3.1) {Energy\\(kWh)};
        \node[draw=cM!50, fill=cM!7, rounded corners=2pt,
              minimum width=1.4cm, minimum height=0.7cm,
              align=center, font=\sffamily\normalsize] at (16.4, 3.1) {CO$_2$\\Emissions};
        \node[draw=cM!50, fill=cM!7, rounded corners=2pt,
              minimum width=1.4cm, minimum height=0.7cm,
              align=center, font=\sffamily\normalsize] at (18.1, 3.1) {Training\\ Duration};

        \draw[SA] (cc.south) -- ++(0,-0.26) -| (13.0, 3.1+0.35);
        \draw[SA] (cc.south) -- ++(0,-0.26) -| (14.7, 3.1+0.35);
        \draw[SA] (cc.south) -- ++(0,-0.26) -| (16.4, 3.1+0.35);
        \draw[SA] (cc.south) -- ++(0,-0.26) -| (18.1, 3.1+0.35);

        \node[draw=cP!70, fill=cP!15, rounded corners=4pt,
              minimum width=5.4cm, minimum height=1.1cm, align=center,
              blur shadow={shadow blur steps=4,shadow xshift=1pt,shadow yshift=-1pt}]
            at (23.0, 6.0) {\textbf{Phase 4}\\[1pt]Performance Evaluation};

        \node[draw=cP!60, fill=cP!10, rounded corners=3pt,
              minimum width=5.0cm, minimum height=0.95cm,
              align=center, inner sep=5pt, font=\sffamily\small] (perf)
            at (23.0, 4.6) {\textbf{Classification Metrics}};

        \node[draw=cP!45, fill=cP!7, rounded corners=2pt,
              minimum width=1.4cm, minimum height=0.7cm,
              align=center, font=\sffamily\scriptsize] at (21.4, 3.1) {Top-$k$\\Accuracy};
        \node[draw=cP!45, fill=cP!7, rounded corners=2pt,
              minimum width=1.4cm, minimum height=0.7cm,
              align=center, font=\sffamily\scriptsize] at (23.0, 3.1) {F$_1$-Score};
        \node[draw=cP!45, fill=cP!7, rounded corners=2pt,
              minimum width=1.4cm, minimum height=0.7cm,
              align=center, font=\sffamily\scriptsize] at (24.6, 3.1) {Acc./kWh\\(Eff.)};

        \draw[SA] (perf.south) -- ++(0,-0.26) -| (21.4, 3.1+0.35);
        \draw[SA] (perf.south) -- ++(0,-0.35) -- (23.0, 3.1+0.35);
        \draw[SA] (perf.south) -- ++(0,-0.26) -| (24.6, 3.1+0.35);

        \draw[MA] (3.3, 4.15) -- node[above, font=\sffamily\scriptsize\bfseries, yshift=1pt]
            {run} (3.7, 4.15);
        \draw[MA] (11.4, 4.15) -- node[above, font=\sffamily\scriptsize\bfseries, yshift=1pt]
            {monitor} (11.8, 4.15);
        \draw[MA] (19.6, 4.15) -- node[above, font=\sffamily\scriptsize\bfseries, yshift=1pt]
            {evaluate} (20.0, 4.15);

        \node[draw=cA!70, fill=cA!15, rounded corners=4pt,
              minimum width=10.0cm, minimum height=1.0cm, align=center,
              blur shadow={shadow blur steps=4,shadow xshift=1pt,shadow yshift=-1pt}] (ph5title)
            at (11.6, 0.6) {\textbf{Phase 5}\\[1pt]\textcolor{black}{Allocation of Variation --- $2^2$ Full Factorial Design}};

        \node[draw=cA!50, fill=cA!7, rounded corners=4pt,
              minimum width=20.4cm, minimum height=0.2cm,
              align=center, inner sep=8pt, font=\sffamily\Large] (anova)
            at (11.6, -0.7) {
            $SS_A = \frac{q_A^2}{n}$\qquad
            $SS_B = \frac{q_B^2}{n}$\qquad
            $SS_{AB} = \frac{q_{AB}^2}{n}$\qquad
            $\mathrm{Influence}_{A} = \frac{SS_A}{SS_T}$\qquad
            $\mathrm{Influence}_{B} = \frac{SS_B}{SS_T}$\qquad
            $\mathrm{Influence}_{AB} = \frac{SS_{AB}}{SS_T}$
        };

        \node[draw=cA!55, fill=cA!8, rounded corners=3pt,
              minimum width=4.5cm, minimum height=1.6cm,
              align=center, inner sep=5pt, font=\sffamily\normalsize]
            at (3.5, -2.7) {\textbf{GPU Power (W)}\\[3pt]};

        \node[draw=cA!55, fill=cA!8, rounded corners=3pt,
              minimum width=4.5cm, minimum height=1.6cm,
              align=center, inner sep=5pt, font=\sffamily\normalsize]
            at (9.8, -2.7) {\textbf{Energy (kWh)}\\[3pt]};

        \node[draw=cA!55, fill=cA!8, rounded corners=3pt,
              minimum width=4.5cm, minimum height=1.6cm,
              align=center, inner sep=5pt, font=\sffamily\normalsize]
            at (16.2, -2.7) {\textbf{CO$_2$ Emissions}\\[3pt]};

        \node[draw=cDk!40, fill=cGr, rounded corners=3pt,
              minimum width=3.2cm, minimum height=1.6cm,
              align=center, inner sep=5pt, font=\sffamily\normalsize]
            at (22.5, -2.7) {
            \textbf{Outputs}\\[2pt]
            Main Effects\\
            Response Values
        };

        \draw[SA] (anova.south) -- ++(0,-0.3) -| ( 3.5, -2.7+0.8);
        \draw[SA] (anova.south) -- ++(0,-0.3) -| ( 9.8, -2.7+0.8);
        \draw[SA] (anova.south) -- ++(0,-0.3) -| (16.2, -2.7+0.8);
        \draw[SA] (anova.south) -- ++(0,-0.3) -| (22.5, -2.7+0.8);

        \draw[MA] (7.55, 1.8) -- (7.55, 1.2);
        \draw[MA] (15.7, 1.8) -- (15.7, 1.2);
        \draw[DA] (23.0, 1.8) -- (23.0, 1.2);

        \foreach \x/\n/\col/\lbl in {
            0.25/1/cF/Testbed,
            7.55/2/cL/Design,
            15.7/3/cM/Monitoring,
            23.0/4/cP/Evaluation}{
          \node[circle, draw=\col, fill=\col!18,
                minimum size=0.8cm, font=\sffamily\bfseries\large, inner sep=0]
              (seq\n) at (\x, 8.0) {\n};
          \node[font=\sffamily\large, text=cDk] at (\x, 7.3) {\lbl};
        }
        
        \node[circle, draw=cA, fill=cA!18,
              minimum size=0.8cm, font=\sffamily\bfseries\large, inner sep=0]
            (seq5) at (26.8, 0.6) {5};
        \node[font=\sffamily\large, text=cDk] at (26.8, -0.1) {Analysis};

        \draw[-{Latex[length=5pt,width=3pt]}, line width=1.2pt, draw=cDk!35]
            (seq1.east) -- (seq2.west);
        \draw[-{Latex[length=5pt,width=3pt]}, line width=1.2pt, draw=cDk!35]
            (seq2.east) -- (seq3.west);
        \draw[-{Latex[length=5pt,width=3pt]}, line width=1.2pt, draw=cDk!35]
            (seq3.east) -- (seq4.west);
        \draw[-{Latex[length=5pt,width=3pt]}, line width=1.2pt, draw=cDk!35]
    (seq4.east) to [out=-2, in=90, looseness=1.4] (seq5.north);

        \end{tikzpicture}
    }
    \caption{Experimental Methodology Workflow.}
    \label{fig:methodology}
\end{figure*}
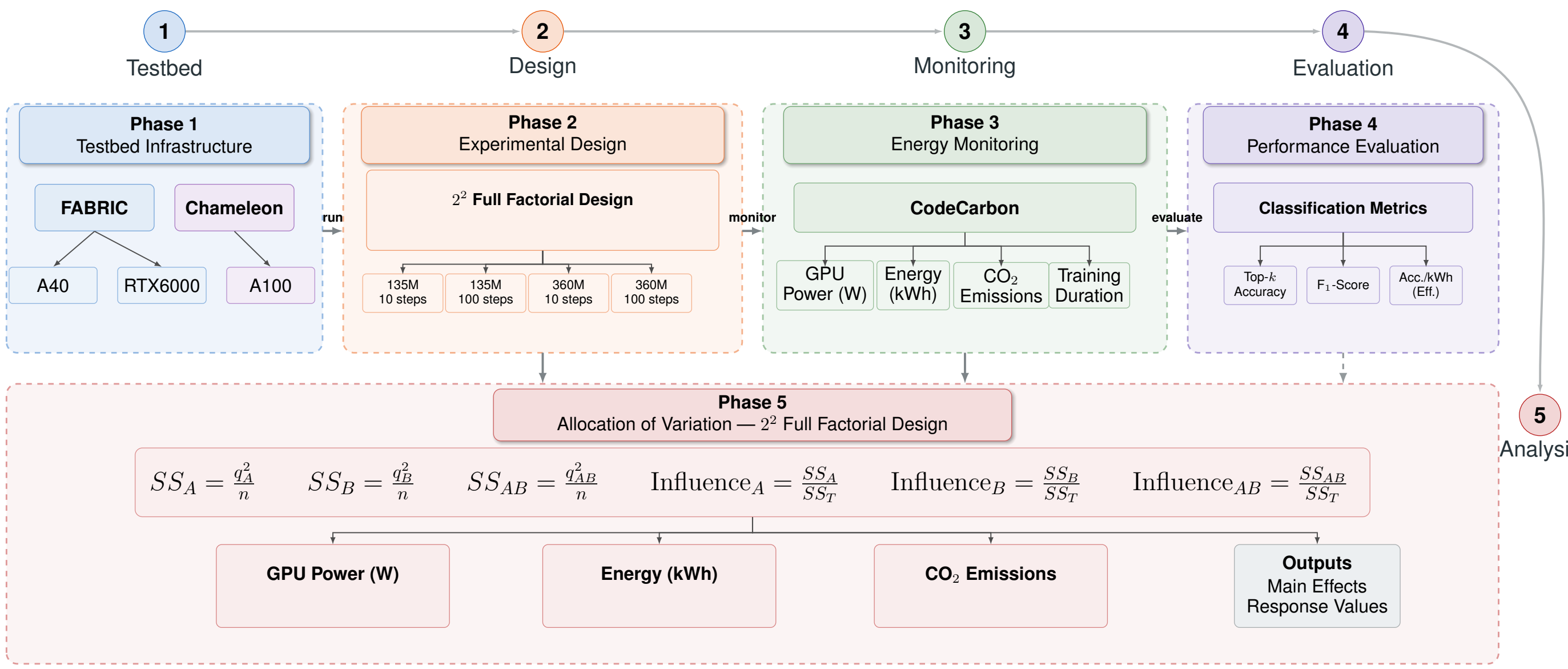

Our method evaluates energy consumption, CO$_2$-equivalent emissions, and predictive performance during federated \ac{LLM} fine-tuning. Figure~\ref{fig:methodology} summarizes five phases: testbed provisioning, experimental design, energy monitoring, performance evaluation, and factorial analysis.

\subsection{Experimental Design}

We used a $2^2$ full factorial design to separate the effects of model size, local optimization intensity, and their interaction~\cite{Jain1991}. The SmolLM-135M and SmolLM-360M models~\cite{Allal2025} were evaluated with 10 and 100 local steps per round, yielding four configurations per \ac{GPU} (Table~\ref{tab:factorial_design}). Both models belong to the same family, limiting architectural differences in the comparison.

\begin{table}[!ht]\centering
\caption{Experimental factors and levels.}\label{tab:factorial_design}
\small
\begin{tabular}{@{}lcc@{}}
\hline
\textbf{Factor} & \textbf{Low level} & \textbf{High level} \\
\hline
$A$, model size & 135M & 360M \\
$B$, local steps per round & 10 & 100 \\
\hline
\end{tabular}
\end{table}

Fine-tuning kept the pretrained backbone frozen and learned \ac{LoRA} updates of the form $\Delta W=BA$, with $B\in\mathbb{R}^{d\times r}$ and $A\in\mathbb{R}^{r\times k}$~\cite{Zanella2024}. The adapter rank was fixed at $r=8$ across all configurations, following the common practice for models of this scale~\cite{3_Hanindhito2025}.

Each experiment comprised 50 federated rounds with 50 simulated clients \textcolor{black}{on one GPU}. At every round, five clients (10\%) were sampled uniformly using a fixed random seed. The selected clients performed the prescribed local steps with a cosine-decayed learning rate. The server then averaged the corresponding adapter parameters with equal client weights and redistributed the global adapter for the next round. \textcolor{black}{Communication energy was unmeasured; aggregation was included in the monitored simulation.}

\subsection{Energy Monitoring}\label{subsec:energy_monitoring}

CodeCarbon monitored each simulation from initialization through the final federated round, covering model loading, client selection, local training, aggregation, and checkpointing~\cite{codecarbon_2025_18129798}. \textcolor{black}{We recorded duration, energy, estimated CO$_2$-equivalent emissions, and GPU power from the NVIDIA Management Library (NVML). Host contributions to CodeCarbon energy remain unresolved, limiting accelerator-specific comparisons. For power $P(t)$ in watts at a consistent accounting boundary and time in seconds, energy follows Eq.~\ref{eq:energy_kwh}:}

\begin{equation}\label{eq:energy_kwh}
E_{\text{kWh}}=\frac{1}{3.6\times10^6}\int_{t_0}^{t_f}P(t)\,dt.
\end{equation}

Emissions were estimated as $E_{\text{kWh}}c_{\text{grid}}$, where $c_{\text{grid}}$ is the location-based average grid carbon intensity in kg CO$_2$e/kWh. \textcolor{black}{Location averages omit intraday variations. Fixed within-platform grid intensity makes energy and emissions contributions identical by construction. Wall-plug validation and resolved CPU/memory accounting are needed to strengthen these software estimates.}

\subsection{Performance Evaluation}

Each round's saved global adapter was evaluated on a held-out HDFS log test set~\cite{zhu2023}. The model predicted the next token at each position in the tokenized sequences. Top-$k$ accuracy measures the fraction of ground-truth next tokens appearing among the $k$ highest-ranked predictions, for $k\in\{1,3,5,10\}$.

For anomaly detection, a sequence was classified as anomalous when its mean Top-$k$ accuracy fell below a threshold $\tau$. \textcolor{black}{We selected $\tau$ from 1,000 uniform values in $[0,1]$ by maximizing $F_1$ on the same test set. Reported $F_1$, precision, and recall carry optimistic selection bias; independent assessment requires validation-only threshold selection.} \textcolor{black}{$F_1$ follows Eq.~\ref{eq:f1_score}:}

\begin{equation}\label{eq:f1_score}
F_1=2\frac{\mathrm{Precision}\,\mathrm{Recall}}{\mathrm{Precision}+\mathrm{Recall}}.
\end{equation}

To relate the predictive performance to the energy expenditure~\cite{8_Kaushik2025}, we also computed according to Eq.~\ref{eq:acc_kwh}:

\begin{equation}\label{eq:acc_kwh}
\mathrm{Efficiency}_k=\frac{\text{Top-}k\text{ accuracy}}{E_{\text{kWh}}}.
\end{equation}

The consolidated comparisons use Top-1 accuracy per kWh. This metric measures energy efficiency, and differences in grid carbon intensity are reflected separately in the emissions estimates.

\subsection{\textcolor{black}{Allocation of Variation}}

\textcolor{black}{For each \ac{GPU}, we allocated observed variation in power, energy, emissions, and duration using the $2^2$ factorial contrasts.} Low and high factor levels were coded as $-1$ and $+1$, with interaction codes $x_{AB,i}=x_{A,i}x_{B,i}$. For treatment responses $y_i$, the signed contrasts are $q_j=\sum_{i=1}^{n}x_{j,i}y_i$, representing model size ($A$), local steps ($B$), and their interaction ($AB$). Their sums of squares \textcolor{black}{follow} Eq.~\ref{eq:4}:

\begin{equation}
\label{eq:4}
SS_j=\frac{q_j^2}{n},\qquad j\in\{A,B,AB\},
\end{equation}
where $n$ is the number of treatment observations. The contribution of each component to the total sum of squares $SS_T$ is according to Eq.~\ref{eq:5}:
\begin{equation}
\label{eq:5}
\mathrm{Influence}_j=\frac{SS_j}{SS_T},\qquad j\in\{A,B,AB\}.
\end{equation}
Multiplying these ratios by 100 yielded the percentage contributions reported in the results. \textcolor{black}{One observation per treatment leaves zero residual degrees of freedom; these descriptive contributions provide neither uncertainty estimates nor significance tests.}

\subsection{Testbed Infrastructure}\label{subsec:testbed_infrastructure}

We evaluated the same training configurations on NVIDIA A40 and RTX\,6000 Ada \acp{GPU} on FABRIC~\cite{Baldin2020FABRIC} and NVIDIA A100 80\,GB PCIe \acp{GPU} on Chameleon Cloud~\cite{Keahey2020Chameleon}. \textcolor{black}{Comparisons combine accelerator, host, and testbed differences, limiting attribution to GPU architecture alone.}

\section{Results and Discussion}\label{sec:results_and_discussion}

\subsection{Energy and local optimization}

\begin{figure*}[!t]
\centering
\begin{tabular}{cc}
    \includegraphics[width=0.49\textwidth]{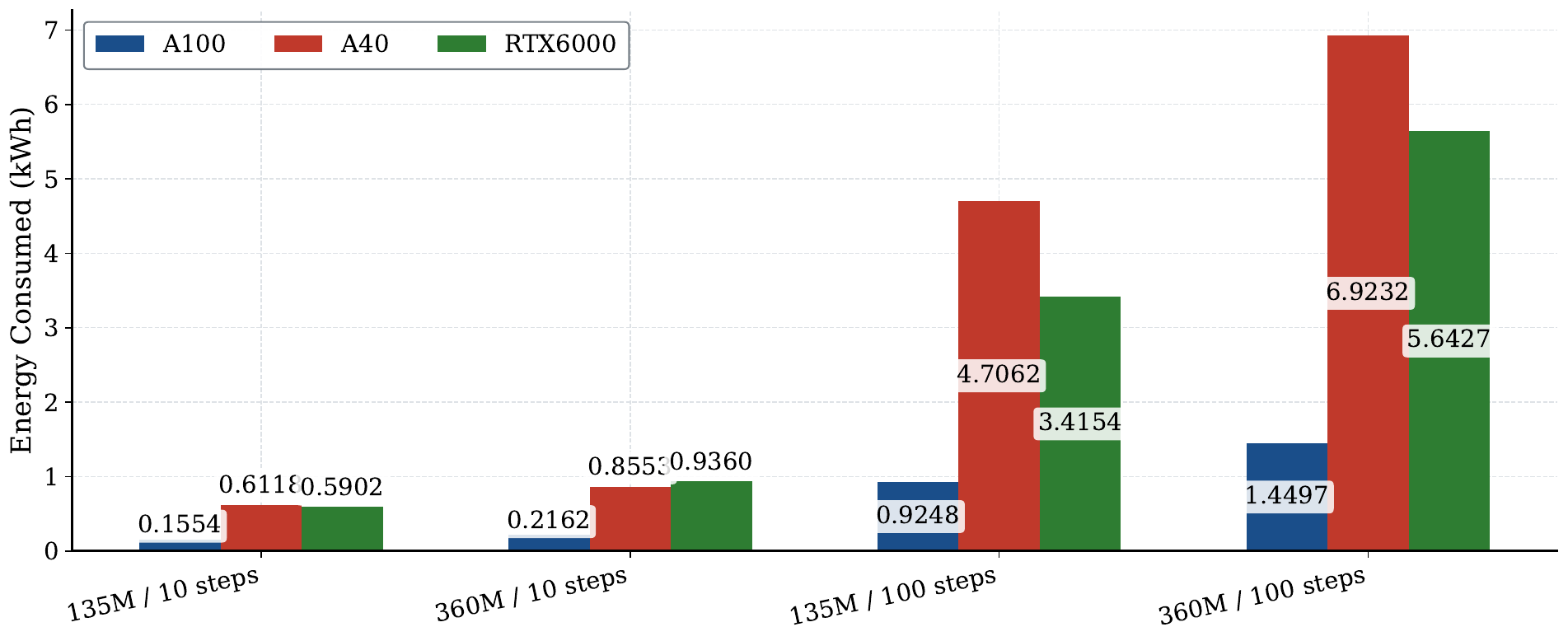} &
    \includegraphics[width=0.49\textwidth]{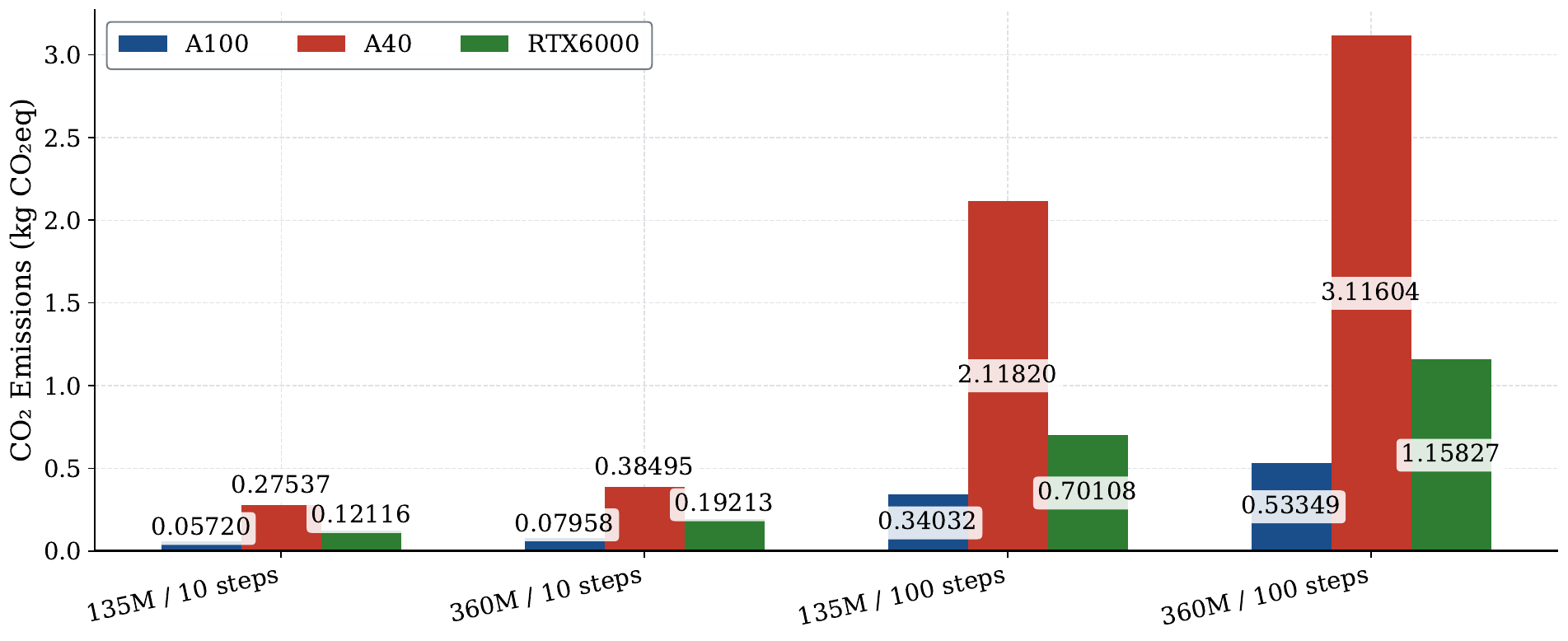} \\
    {\footnotesize (a) Energy consumption} &
    {\footnotesize (b) CO$_2$-equivalent emissions}
\end{tabular}
\caption{Energy consumption and CO$_2$-equivalent emissions by device and configuration.}
\label{fig:energy_emissions}
\end{figure*}

Increasing local optimization from 10 to 100 steps per round raised energy consumption by \textcolor{black}{approximately} 6--8$\times$ across the tested devices and model sizes according to Fig.~\ref{fig:energy_emissions}. With 10 steps, all configurations remained below 1\,kWh. With 100 steps, consumption exceeded 6.9\,kWh on A40 and 5.6\,kWh on RTX\,6000 Ada, while A100 stayed below 1.5\,kWh. For the 360M model with 100 steps, CO$_2$-equivalent emissions reached 3.12, 1.16, and 0.53\,kg on A40, RTX\,6000 Ada, and A100, respectively.

\begin{figure}[!t]
\centering
\includegraphics[width=\columnwidth]{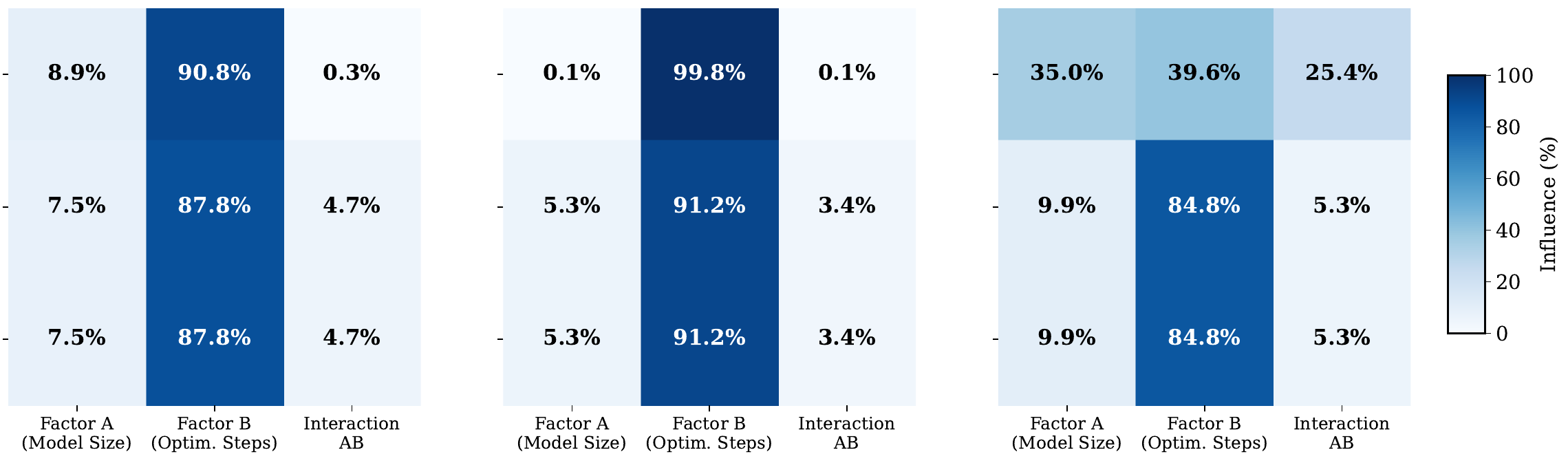}
\caption{Factor contributions to each \textcolor{black}{recorded} response.}
\label{fig:factorial_heatmap}
\end{figure}

\textcolor{black}{Figure~\ref{fig:factorial_heatmap} allocates the observed variation.} Local steps accounted for 91.2\%, 87.8\%, and 84.8\% of the \textcolor{black}{observed variation} in energy and emissions on A40, A100, and RTX\,6000 Ada, respectively. Model size contributed less than 10\% \textcolor{black}{for energy and emissions}, and the interaction accounted for at most 5.3\%. \textcolor{black}{This ordering is conditional on the unequal proportional ranges chosen for steps and model size.}

Model size nevertheless increased the cost of longer training. On A40, the energy gap between the 135M and 360M models grew from 0.24\,kWh at 10 steps to 2.22\,kWh at 100 steps; on RTX\,6000 Ada, it grew from 0.35 to 2.23\,kWh. Interaction contributions of 3.4--5.3\% indicate that larger models and longer optimization compound the energy cost. \textcolor{black}{Intermediate step counts and nonlinear responses remain outside this two-level design.}

\begin{figure*}[!t]
\centering
\begin{tabular}{cc}
    \includegraphics[width=0.45\textwidth]{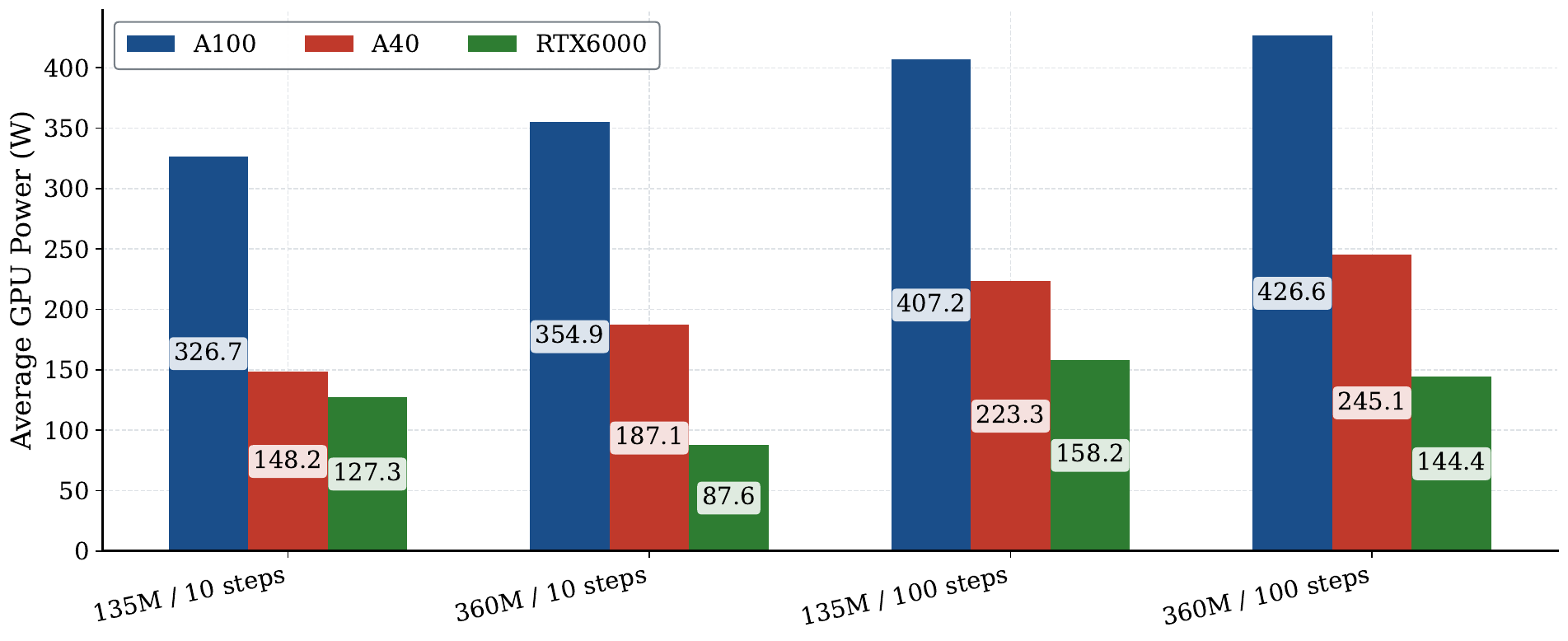} &
    \includegraphics[width=0.45\textwidth]{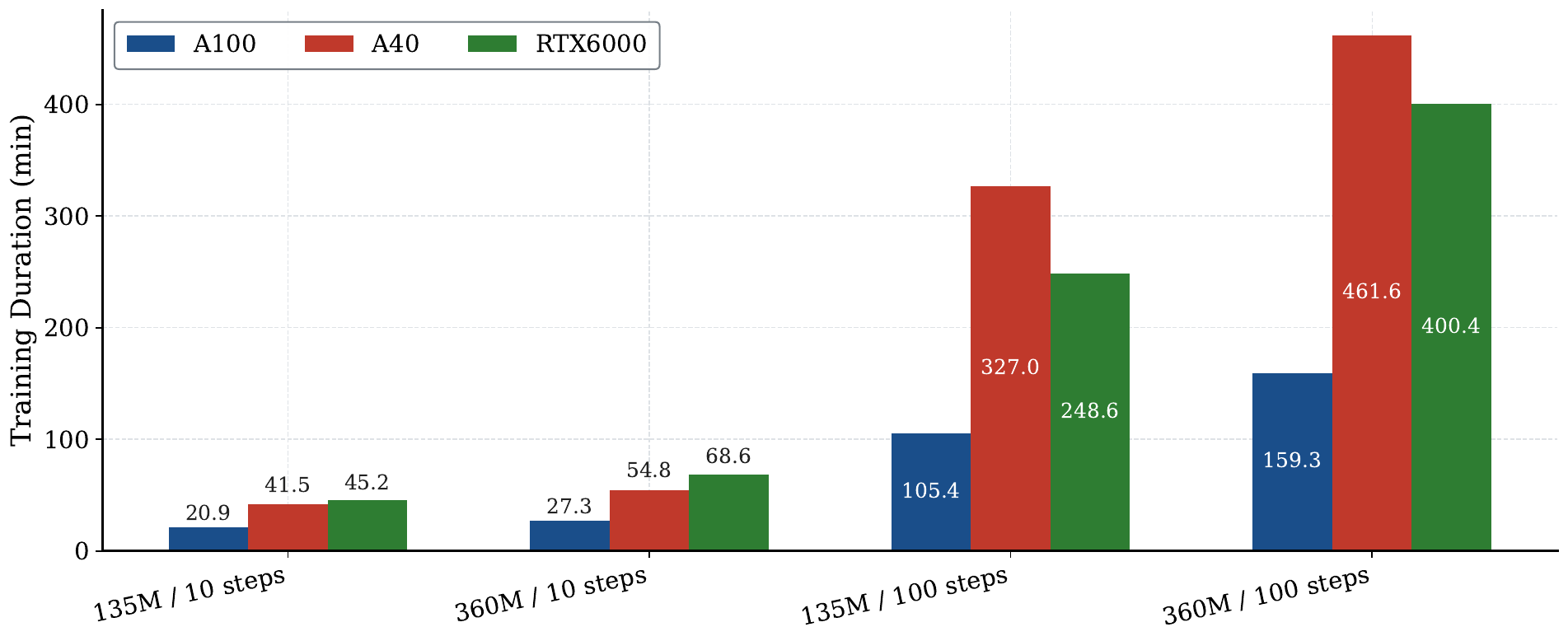} \\
    {\footnotesize (a) Average \textcolor{black}{GPU} power} &
    {\footnotesize (b) Training duration}
\end{tabular}
\caption{Average \textcolor{black}{GPU} power and training duration by device and configuration.}
\label{fig:power_duration}
\end{figure*}

\subsection{Hardware and training duration}

\textcolor{black}{Figure~\ref{fig:power_duration}(a) reports average GPU power separately from CodeCarbon energy. Reconciling their accounting boundaries requires component-level logs.} For this power metric, model size, steps, and their interaction explained 35.0\%, 39.6\%, and 25.4\% of the variance on RTX\,6000 Ada, while steps alone explained 99.8\% on A40 according to Fig.~\ref{fig:factorial_heatmap}. \textcolor{black}{The RTX non-monotonic response may reflect run variability; replication is needed to assess it.}

\textcolor{black}{The recorded energy costs combine accelerator, host, and duration effects.} Training duration helps explain this result according to Fig.~\ref{fig:power_duration}(b). With 100 steps, A40 and RTX\,6000 Ada required 2--3$\times$ the wall-clock time of A100, accumulating energy over longer runs. \textcolor{black}{Unresolved host accounting and A100 predictive quality restrict hardware comparisons to these observations. Efficiency rankings require component-level energy and matched quality.}

\begin{figure*}[!t]
\centering
\begin{tabular}{c}
    \includegraphics[width=0.75\textwidth]{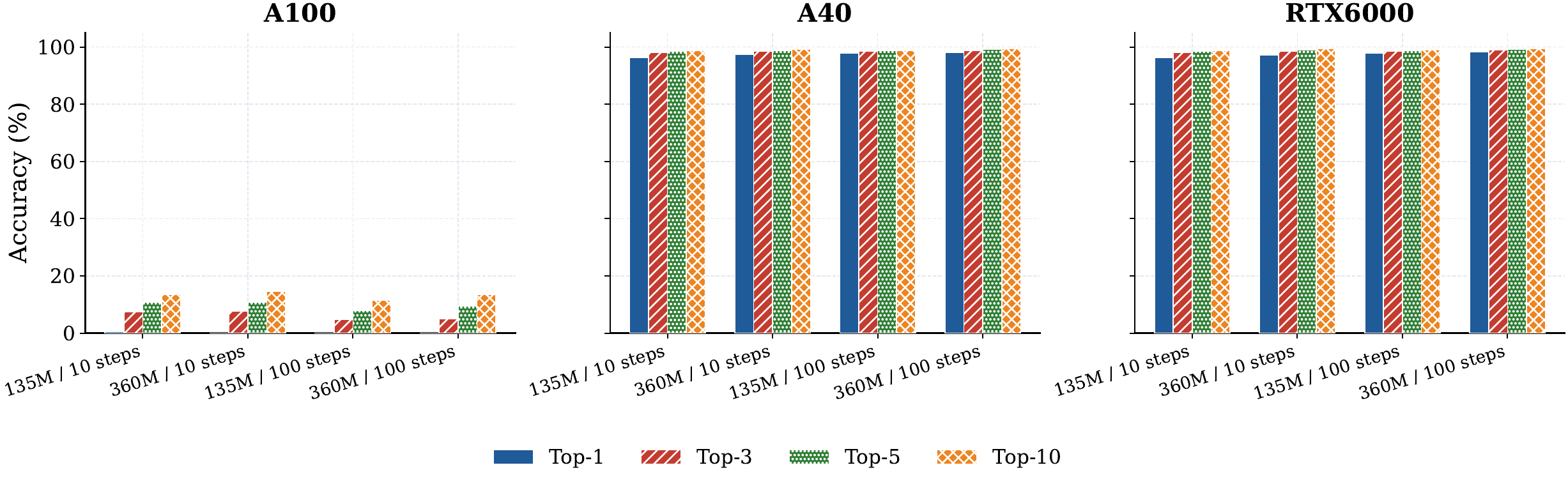} \\
    {\footnotesize (a) Top-$k$ accuracy \textcolor{black}{profiles}} \\[10pt] % Espaço vertical entre (a) e a imagem (b)
    \includegraphics[width=0.75\textwidth]{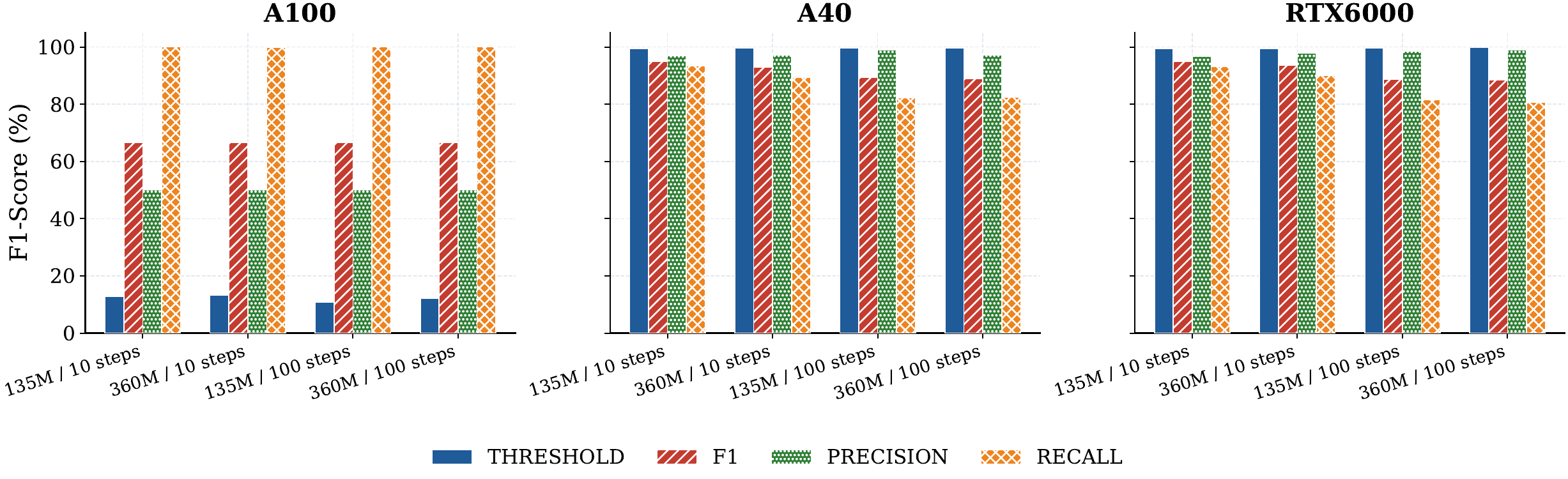} \\
    {\footnotesize (b) F$_1$, precision, and recall by configuration}
\end{tabular}
\caption{Model performance profiles during training and evaluation.}
\label{fig:accuracy_metrics}
\end{figure*}

\subsection{Predictive performance and energy efficiency}

A40 and RTX\,6000 Ada achieved near-ceiling Top-1 accuracy and high F$_1$ scores across all four configurations (Fig.~\ref{fig:accuracy_metrics}). \textcolor{black}{The F$_1$ scores retain the threshold-selection bias described above.} Increasing local optimization to 100 steps provided little additional accuracy despite its higher energy cost. \textcolor{black}{These results favor 10 steps for this HDFS workload.}

\begin{figure*}[!t]
\centering
\begin{tabular}{cc}
    \includegraphics[width=0.49\textwidth]{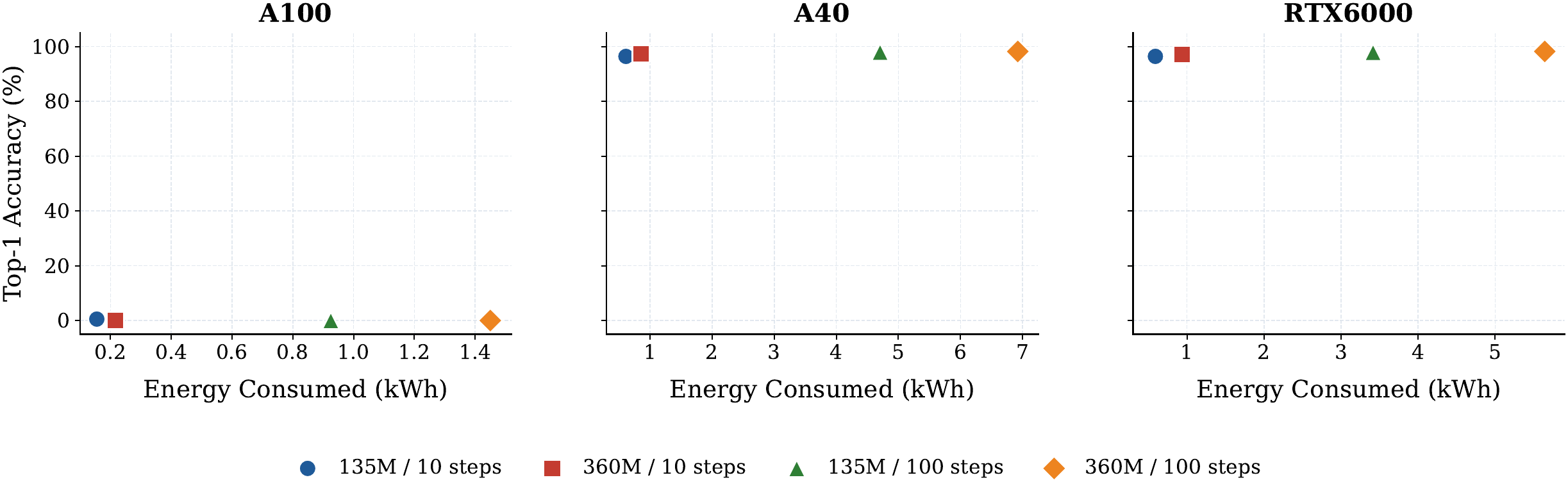} &
    \includegraphics[width=0.49\textwidth]{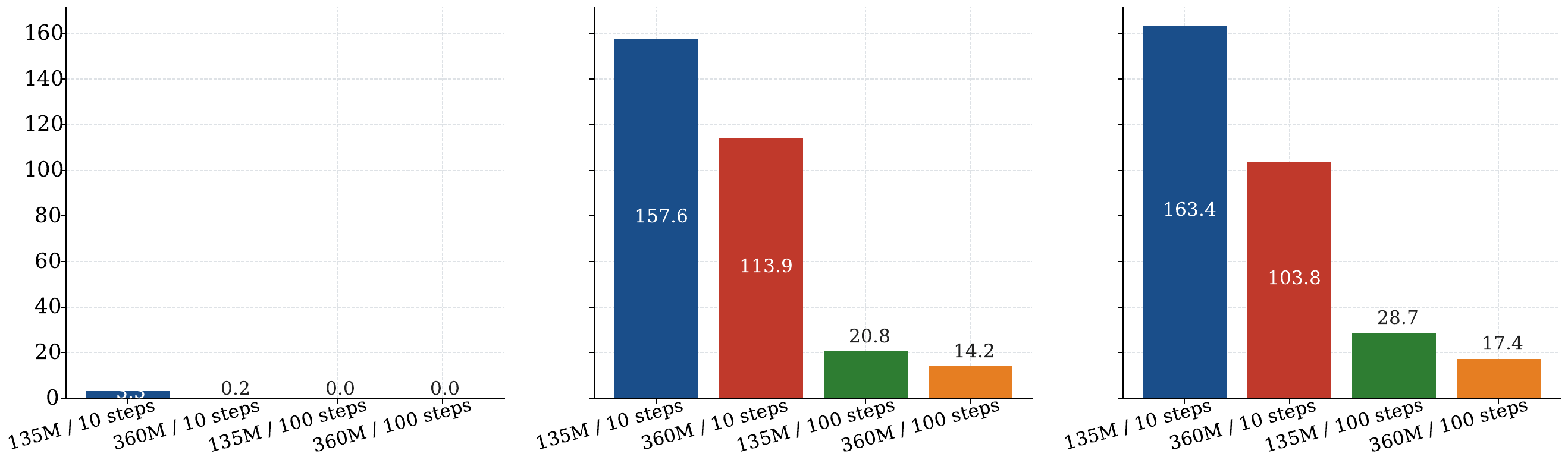} \\
    {\footnotesize (a) Top-1 accuracy versus energy consumption} &
    {\footnotesize (b) Top-1 accuracy per kWh}
\end{tabular}
\caption{Accuracy--energy trade-off and energy efficiency by device and configuration.}
\label{fig:accuracy_efficiency}
\end{figure*}

The accuracy--energy comparison in Fig.~\ref{fig:accuracy_efficiency}(a) makes this trade-off clear: on A40 and RTX\,6000 Ada, 10-step configurations combined high accuracy with low energy consumption, whereas 100 steps mainly increased the energy cost. Accuracy per kWh likewise favored shorter training as in Fig.~\ref{fig:accuracy_efficiency}(b). \textcolor{black}{These observations motivate adaptive local-step budgets for this workload.} A100 showed lower accuracy and F$_1$ across all configurations despite its lower \textcolor{black}{recorded} energy consumption. \textcolor{black}{Hardware-efficiency comparisons require checking checkpoint, tokenizer, precision, and evaluation consistency, then assessing energy at matched predictive quality.}

\FloatBarrier

%\subsection{Scope and limitations}
%All clients were simulated on a single \ac{GPU} per run, assuming uniform compute capability. Real federated deployments distribute this workload across heterogeneous devices and incur communication costs, which were not measured here. Moreover, without centralized or full fine-tuning baselines, the reported results do not isolate the energy savings attributable specifically to federation or to \ac{LoRA}. The findings therefore describe configuration choices within the evaluated federated pipeline.

\section{Concluding Remarks}\label{sec:concluding_remarks}
 
\textcolor{black}{In this paper, we evaluated energy consumption and estimated carbon emissions during federated \ac{LoRA} fine-tuning. Within the tested ranges, local optimization dominated their observed variation. Smaller local budgets retained comparable accuracy on A40 and RTX\,6000 Ada. The evaluation covers one model family, dataset, and seed; platform comparisons depend on accounting boundaries and predictive quality. Future work should examine independent threshold selection, replicated runs, communication energy, and adaptive local budgets on heterogeneous consumer devices.}

\section*{Acknowledgment}

%Omitted due to double-blind review process.

The authors acknowledge support from FAPEMIG under Grants \#APQ00923-24 and \#APQ-00769-26; the FAPESP MCTIC/CGI Research Project SFI2 -- Slicing Future Internet Infrastructures under Grant 2018/23097-3; and Fundação para a Ciência e a Tecnologia (FCT) through the R\&D Unit Project UID/00319/2025, Centro ALGORITMI (ALGORITMI/UM) and the ANIMA Institute for supporting this work.

\bibliographystyle{IEEEtran}
\bibliography{example}

\begin{thebibliography}{10}
\providecommand{\url}[1]{#1}
\csname url@samestyle\endcsname
\providecommand{\newblock}{\relax}
\providecommand{\bibinfo}[2]{#2}
\providecommand{\BIBentrySTDinterwordspacing}{\spaceskip=0pt\relax}
\providecommand{\BIBentryALTinterwordstretchfactor}{4}
\providecommand{\BIBentryALTinterwordspacing}{\spaceskip=\fontdimen2\font plus
\BIBentryALTinterwordstretchfactor\fontdimen3\font minus \fontdimen4\font\relax}
\providecommand{\BIBforeignlanguage}[2]{{%
\expandafter\ifx\csname l@#1\endcsname\relax
\typeout{** WARNING: IEEEtran.bst: No hyphenation pattern has been}%
\typeout{** loaded for the language `#1'. Using the pattern for}%
\typeout{** the default language instead.}%
\else
\language=\csname l@#1\endcsname
\fi
#2}}
\providecommand{\BIBdecl}{\relax}
\BIBdecl

\bibitem{su18031359}
\BIBentryALTinterwordspacing
A.~P. Oliveira, T.~Carraquico, and C.~Martinez-Perez, ``Beyond efficiency: A systematic review of energy consumption and carbon footprint across the ai lifecycle,'' \emph{Sustainability}, vol.~18, no.~3, 2026. [Online]. Available: \url{https://www.mdpi.com/2071-1050/18/3/1359}
\BIBentrySTDinterwordspacing

\bibitem{Alexandre2026}
\BIBentryALTinterwordspacing
A.~d'Orgeval, S.~Sheehan, Q.~Avenas, E.~Assoumou, and V.~Sessa, ``Generative ai impact assessment through a life cycle analysis of multiple data center typologies,'' \emph{Applied Energy}, vol. 406, p. 127288, 2026. [Online]. Available: \url{https://www.sciencedirect.com/science/article/pii/S0306261925020185}
\BIBentrySTDinterwordspacing

\bibitem{goldmansachs2024_ai_power_demand}
{Goldman Sachs Research}, ``Ai is poised to drive 160\% increase in data center power demand,'' \url{https://www.goldmansachs.com/insights/articles/AI-poised-to-drive-160-increase-in-power-demand}, May 2024, accessed: 2026-03-02.

\bibitem{Wu2024}
\BIBentryALTinterwordspacing
F.~Wu, Z.~Li, Y.~Li, B.~Ding, and J.~Gao, ``Fedbiot: Llm local fine-tuning in federated learning without full model,'' in \emph{Proceedings of the 30th ACM SIGKDD Conference on Knowledge Discovery and Data Mining}, ser. KDD '24.\hskip 1em plus 0.5em minus 0.4em\relax New York, NY, USA: Association for Computing Machinery, 2024, p. 3345–3355. [Online]. Available: \url{https://doi.org/10.1145/3637528.3671897}
\BIBentrySTDinterwordspacing

\bibitem{Hinov2025}
\BIBentryALTinterwordspacing
N.~Hinov, ``The energy hunger of ai: Large language models as challenges and enablers for sustainable energy,'' \emph{Energies}, vol.~18, no.~17, 2025. [Online]. Available: \url{https://www.mdpi.com/1996-1073/18/17/4701}
\BIBentrySTDinterwordspacing

\bibitem{Moreira2026}
L.~F. Rodrigues~Moreira, R.~Moreira, and L.~G. Ferreira~Rodrigues, ``Energy-aware ensemble learning for coffee leaf disease classification,'' in \emph{2026 IEEE 23rd Consumer Communications \& Networking Conference (CCNC)}, 2026, pp. 1--6.

\bibitem{1_Iftikhar2025}
S.~Iftikhar, S.~H. Alsamhi, and S.~Davy, ``Enhancing sustainability in llm training: Leveraging federated learning and parameter-efficient fine-tuning,'' \emph{IEEE Transactions on Sustainable Computing}, vol.~10, no.~6, pp. 1158--1172, 2025.

\bibitem{13_Herbert2024}
\BIBentryALTinterwordspacing
H.~Woisetschl\"{a}ger, A.~Erben, S.~Wang, R.~Mayer, and H.-A. Jacobsen, ``Federated fine-tuning of llms on the very edge: The good, the bad, the ugly,'' in \emph{Proceedings of the Eighth Workshop on Data Management for End-to-End Machine Learning}, ser. DEEM '24.\hskip 1em plus 0.5em minus 0.4em\relax New York, NY, USA: Association for Computing Machinery, 2024, p. 39–50. [Online]. Available: \url{https://doi.org/10.1145/3650203.3663331}
\BIBentrySTDinterwordspacing

\bibitem{4_Wu2026}
F.~Wu, J.~Hu, G.~Min, and S.~Wang, ``Adaptive rank allocation for federated parameter-efficient fine-tuning of language models,'' \emph{IEEE Transactions on Computers}, pp. 1--14, 2026.

\bibitem{7_Thuau2025}
S.~Thuau, S.~Haidar, A.~Bajracharya, and R.~Chelouah, ``Frugal federated learning for violence detection: A comparison of lora-tuned vlms and personalized cnns,'' in \emph{2025 3rd International Conference on Federated Learning Technologies and Applications (FLTA)}, 2025, pp. 196--202.

\bibitem{10_Xu2024}
\BIBentryALTinterwordspacing
M.~Xu, D.~Cai, Y.~Wu, X.~Li, and S.~Wang, ``{FwdLLM}: Efficient federated finetuning of large language models with perturbed inferences,'' in \emph{2024 USENIX Annual Technical Conference (USENIX ATC 24)}.\hskip 1em plus 0.5em minus 0.4em\relax Santa Clara, CA: USENIX Association, Jul. 2024, pp. 579--596. [Online]. Available: \url{https://www.usenix.org/conference/atc24/presentation/xu-mengwei}
\BIBentrySTDinterwordspacing

\bibitem{2_Li2025}
Z.~Li, S.~Wu, L.~Li, and S.~Zhang, ``Energy-efficient split learning for fine-tuning large language models in edge networks,'' \emph{IEEE Networking Letters}, vol.~7, no.~3, pp. 176--180, 2025.

\bibitem{12_Zhao2026}
Y.~Zhao, L.~Feng, F.~Zhou, W.~Li, Z.~Xiong, H.~Du, C.~Wu, S.~Guo, T.~Q.~S. Quek, and Z.~Han, ``Energy-efficient over-the-air computation for federated generative model fine-tuning in unmanned vehicle-assisted disaster relief,'' \emph{IEEE Transactions on Cognitive Communications and Networking}, vol.~12, pp. 56--73, 2026.

\bibitem{9_Krzywaniak2023}
\BIBentryALTinterwordspacing
A.~Krzywaniak, P.~Czarnul, and J.~Proficz, ``Dynamic gpu power capping with online performance tracing for energy efficient gpu computing using depo tool,'' \emph{Future Generation Computer Systems}, vol. 145, pp. 396--414, 2023. [Online]. Available: \url{https://www.sciencedirect.com/science/article/pii/S0167739X23001267}
\BIBentrySTDinterwordspacing

\bibitem{11_Newkirk2025}
\BIBentryALTinterwordspacing
A.~C. Newkirk, J.~Fernandez, J.~Koomey, I.~Latif, E.~Strubell, A.~Shehabi, and C.~Samaras, ``Empirically-calibrated h100 node power models for accurate ai training energy estimation,'' \emph{Environmental Research: Energy}, vol.~2, no.~4, p. 045016, dec 2025. [Online]. Available: \url{https://doi.org/10.1088/2753-3751/ae2486}
\BIBentrySTDinterwordspacing

\bibitem{15_Gogineni2025}
K.~Gogineni, A.~Suvizi, and G.~Venkataramani, ``Llms on a budget: System-level approaches to power-efficient and scalable fine-tuning,'' \emph{IEEE Open Journal of the Computer Society}, vol.~6, pp. 987--1000, 2025.

\bibitem{14_Ozkan2025}
\BIBentryALTinterwordspacing
M.~Ozkan and C.~S. Ozkan, ``Federated carbon intelligence for sustainable ai: Real-time optimization across heterogeneous hardware fleets,'' \emph{MRS Energy {\&} Sustainability}, Nov 2025. [Online]. Available: \url{https://doi.org/10.1557/s43581-025-00146-1}
\BIBentrySTDinterwordspacing

\bibitem{5_Li2025}
\BIBentryALTinterwordspacing
Z.~Li, P.~Tang, X.~Wang, X.~Liu, and P.~Mou, ``Pcf-rwkv: Large language model for product carbon footprint estimation,'' \emph{Sustainability}, vol.~17, no.~3, 2025. [Online]. Available: \url{https://www.mdpi.com/2071-1050/17/3/1321}
\BIBentrySTDinterwordspacing

\bibitem{6_Kuang2024}
\BIBentryALTinterwordspacing
W.~Kuang, B.~Qian, Z.~Li, D.~Chen, D.~Gao, X.~Pan, Y.~Xie, Y.~Li, B.~Ding, and J.~Zhou, ``Federatedscope-llm: A comprehensive package for fine-tuning large language models in federated learning,'' in \emph{Proceedings of the 30th ACM SIGKDD Conference on Knowledge Discovery and Data Mining}, ser. KDD '24.\hskip 1em plus 0.5em minus 0.4em\relax New York, NY, USA: Association for Computing Machinery, 2024, p. 5260–5271. [Online]. Available: \url{https://doi.org/10.1145/3637528.3671573}
\BIBentrySTDinterwordspacing

\bibitem{3_Hanindhito2025}
\BIBentryALTinterwordspacing
B.~Hanindhito, B.~Patel, and L.~K. John, ``Large language model fine-tuning with low-rank adaptation: A performance exploration,'' in \emph{Proceedings of the 16th ACM/SPEC International Conference on Performance Engineering}, ser. ICPE '25.\hskip 1em plus 0.5em minus 0.4em\relax New York, NY, USA: Association for Computing Machinery, 2025, p. 92–104. [Online]. Available: \url{https://doi.org/10.1145/3676151.3719377}
\BIBentrySTDinterwordspacing

\bibitem{8_Kaushik2025}
B.~Kaushik, A.~Taneja, and S.~Dahiya, ``Toward sustainable ai: A review of energy-efficient large language models,'' in \emph{2025 8th International Conference on Computing Methodologies and Communication (ICCMC)}, 2025, pp. 944--951.

\bibitem{Jain1991}
R.~Jain, \emph{The art of computer systems performance analysis}.\hskip 1em plus 0.5em minus 0.4em\relax john wiley \& sons, 1990.

\bibitem{Allal2025}
\BIBentryALTinterwordspacing
L.~B. Allal, A.~Lozhkov, E.~Bakouch, G.~M. Blázquez, G.~Penedo, L.~Tunstall, A.~Marafioti, H.~Kydlíček, A.~P. Lajarín, V.~Srivastav, J.~Lochner, C.~Fahlgren, X.-S. Nguyen, C.~Fourrier, B.~Burtenshaw, H.~Larcher, H.~Zhao, C.~Zakka, M.~Morlon, C.~Raffel, L.~von Werra, and T.~Wolf, ``Smollm2: When smol goes big -- data-centric training of a small language model,'' 2025. [Online]. Available: \url{https://arxiv.org/abs/2502.02737}
\BIBentrySTDinterwordspacing

\bibitem{Zanella2024}
M.~Zanella and I.~Ben~Ayed, ``Low-rank few-shot adaptation of vision-language models,'' in \emph{Proceedings of the IEEE/CVF Conference on Computer Vision and Pattern Recognition (CVPR) Workshops}, June 2024, pp. 1593--1603.

\bibitem{codecarbon_2025_18129798}
\BIBentryALTinterwordspacing
B.~Courty, V.~Schmidt \emph{et~al.}, ``mlco2/codecarbon: v3.2.1,'' Zenodo, 2025. [Online]. Available: \url{https://doi.org/10.5281/zenodo.18129798}
\BIBentrySTDinterwordspacing

\bibitem{zhu2023}
\BIBentryALTinterwordspacing
J.~Zhu, S.~He, P.~He, J.~Liu, and M.~R. Lyu, ``Loghub: A large collection of system log datasets for ai-driven log analytics,'' 2023. [Online]. Available: \url{https://arxiv.org/abs/2008.06448}
\BIBentrySTDinterwordspacing

\bibitem{Baldin2020FABRIC}
I.~Baldin, A.~Nikolich, J.~Griffioen, I.~S. Monga, K.-C. Wang, T.~Lehman, and P.~Ruth, ``Fabric: A national-scale programmable experimental network infrastructure,'' \emph{IEEE Internet Computing}, vol.~24, no.~3, pp. 38--47, 2020.

\bibitem{Keahey2020Chameleon}
K.~Keahey, J.~Anderson, Z.~Zhen, P.~Riteau, P.~Ruth, D.~Stanzione, M.~Cevik, J.~Colleran, H.~S. Gunawi, C.~Hammock, J.~Mambretti, A.~Barnes, F.~Halbach, A.~Rocha, and J.~Stubbs, ``Lessons learned from the chameleon testbed,'' in \emph{USENIX Annual Technical Conference (USENIX ATC '20)}.\hskip 1em plus 0.5em minus 0.4em\relax USENIX Association, Jul. 2020.

\end{thebibliography}

\end{document}